\documentclass[11pt,a4paper]{article}
\usepackage{times,latexsym}
\usepackage[T1]{fontenc}
\usepackage[utf8]{inputenc}
\usepackage{url}
\usepackage{amsmath,amssymb,amsfonts}
\usepackage{graphicx}
\usepackage{booktabs}
\usepackage{array}
\usepackage{float}
\usepackage{microtype}

\usepackage[acceptedWithA]{tacl2021v1}


\title{What Does MMLU Actually Measure? \\ A Psychometric Audit of Difficulty Structure in Aggregate Benchmark Scores}

\author{Dana Paquin \\
  Stanford University \\
  \texttt{dpaquin@stanford.edu} \And
  Riddhiman Jain \\
  \texttt{riddhimanjain15@gmail.com}}

\date{}

\begin{document}
\maketitle

\begin{abstract}
Although MMLU is widely adopted as a benchmark for calibrating general AI capabilities, we psychometrically demonstrate that its aggregate score primarily evaluates a model's factual retrieval capacity rather than its reasoning ability. By calibrating item difficulty for 1,000 open-weights language models over 14,042 MMLU test items using Item Response Theory, we show that evaluating both abilities via a single test is inherently flawed. Difficulty is then regressed on a deterministic, text-extractable framework of structural complexity. Applying a joint Wald test with subject-clustered covariances demonstrates that the MMLU conflates fundamentally separable constructs. The mapping from structural complexity to difficulty is not invariant across the benchmark's STEM and non-STEM partitions. This finding has practical consequences. Aggregate leaderboard ranks track non-STEM accuracy more closely than STEM accuracy, so selecting a Top-50 model on the aggregate for a reasoning-intensive deployment displaces roughly 22\% of the STEM-appropriate choices. Furthermore, when controlling for the multiple-choice guessing floor natively inside the response model, we find that higher-ability models continue to degrade more steeply under increased reasoning depth. The MMLU aggregate therefore weights retrieval capacity and reasoning stability unequally, inadvertently favoring models optimized for retrieval. We release our deterministic framework as a reproducible auditing instrument and recommend disaggregated reporting.
\end{abstract}

\section{Introduction}\label{introduction}

The Massive Multitask Language Understanding (MMLU) benchmark \citep{hendrycks2021} evaluates large language models (LLMs) on 14,042 multiple-choice items across 57 subjects and reports a single aggregate accuracy score, which practitioners widely read as a unified measure of general capability. Whether one score psychometrically captures a coherent construct is an old question: the coherence of a single ability score has been contested since \citet{spearman1904} introduced a general intelligence factor $g$, which \citet{thurstone1938} and \citet{carroll1993} showed dissolves into separable primary abilities under factor analysis. ACT-R draws the distinction that matters here, separating declarative knowledge, the retrieval of facts, from procedural knowledge, the execution of multi-step logical operations \citep{anderson1983,tulving1972}, and we operationalize it directly: Zipf Rarity and Entity Density capture a prompt's declarative retrieval demands, while the Weighted Symbolic Complexity Graph (WSCG) captures its sequential inferential load. If MMLU items draw unequally on these processes, aggregating them into a single metric risks conflating separable abilities.

There is also empirical reason to expect MMLU in particular to be retrieval-weighted. \citet{wang2024} documented that Chain-of-Thought prompting fails to improve, and sometimes degrades, MMLU accuracy while yielding large gains on their successor MMLU-Pro, concluding that MMLU is ``mostly knowledge-driven without requiring too much reasoning.'' That diagnosis was reached heuristically, though. MMLU-Pro filtered on item difficulty ($b_{i}$), not discrimination ($a_{i}$), which narrows the measurement range but does not target items that fail to separate models, and no psychometric test was applied to tell a ``knowledge-driven'' item from a ``reasoning-driven'' one.

To investigate exactly what MMLU psychometrically measures, we conduct an explanatory item-response audit of its difficulty structure. Building on the foundational application of Item Response Theory (IRT) to NLP \citep{lalor2019}, we calibrate item difficulty and model ability with a Two-Parameter Logistic (2PL) model fitted to roughly 14 million responses from 1,000 open-weights models on the Open LLM Leaderboard \citep{beeching2023}, and regress the estimated difficulties on a deterministic, text-extractable framework of five dimensions and nine indicators. We then test whether that mapping is structurally invariant across domains. The partition we test it on comes from the benchmark designers' \emph{a priori} taxonomy, fixed independently of anything we find, and not a boundary we drew ourselves after looking at the data, on the working hypothesis that STEM items generally demand higher procedural reasoning while non-STEM items rely more heavily on factual retrieval. The dichotomy is imperfect, since College Biology (STEM) requires substantial recall whereas Formal Logic (non-STEM) requires inferential chaining, but it is a predefined, unbiased boundary against which homogeneity can fail.

MMLU's two designer-defined partitions turn out not to measure a single interchangeable ability. A confirmatory two-dimensional IRT model, in which each item loads only on the dimension its designer-assigned subject specifies, puts the latent correlation between STEM and non-STEM ability at $\rho = 0.965$, strictly below unity and with no posterior mass above $0.99$, predicts held-out responses better than the unidimensional model, and is corroborated by a disattenuated subscale correlation of $0.977$ at matched test length. The mapping from text structure to item difficulty is also not the same function in the two partitions. Only one contrast, Entity Density, survives subject-clustered covariances, a wild cluster bootstrap and Holm correction over the family of nine, acting as an active constraint inside a STEM item and as a retrieval cue outside one. Downstream, aggregate rank tracks non-STEM accuracy more closely than STEM accuracy, even after the partitions are equalized for test length. We release the framework, the features and the IRT estimates so this audit can be repeated on any benchmark whose items are text.

\section{Related Work}\label{related-work}

Aggregate benchmark scores masking construct conflation is well documented. \citet{bender2021} argue that knowledge-retrieval performance is no evidence of inferential capacity, \citet{mccoy2019} showed BERT-family models relying on surface heuristics rather than logical inference on NLI benchmarks, and high GLUE scores have been catalogued beside near-random performance on reasoning probes \citep{rogers2020}. The strongest statement is \citeauthor{bowman2021}'s (\citeyear{bowman2021}), that benchmarks framed as measures of general ability rarely possess the construct validity such framing presupposes \citep[see also][]{raji2021}. We present psychometric evidence corroborating those critiques. \citet{lalor2019} pioneered IRT for NLP benchmarks, \citet{rodriguez2021} built leaderboards on that observation and \citet{polo2024} extended the line with tinyBenchmarks, but all three assume unidimensionality, i.e.~a single latent ability $\theta_{j}$ per model, and hence cannot detect whether that dimension conflates separable constructs contributing non-uniformly across domains. Exploratory multidimensional IRT \citep{reckase2009} recovers its factors from the response matrix and interprets them post hoc, which makes the dimensions a product of the analysis rather than a hypothesis the data can reject, so we fit a confirmatory model whose loading pattern is fixed by MMLU's own designers (Appendix~\ref{sec:mirt}).

\textbf{What we are and are not testing.} Our procedure is a multi-group test of an explanatory item-difficulty model, i.e.~a Linear Logistic Test Model \citep{fischer1973} extended with an error term and estimated separately in two groups of items \citep{deboeck2004}. It is the item-side analogue of measurement invariance, not invariance itself \citep{meredith1993,vandenberg2000}: the respondent population is fixed, the groups are groups of items, and the null holds constant a regression of item parameters on item covariates. The respondent-side test is separately available to us with model families in place of examinee groups, and Appendix~\ref{sec:labels} reports Mantel--Haenszel differential item functioning across architecture families \citep{holland1988,dorans1993}. The relevant validity is construct representation rather than nomothetic span \citep{embretson1983}.

Predicting item difficulty from text has precedent in educational measurement \citep{embretson1987} and in NLP \citep{vania2021}, while domain-targeted evaluations such as ARC \citep{clark2018} and AGIEval \citep{zhong2023} instead construct item pools isolating particular reasoning types. Our syntactic indicators rest on Dependency Locality Theory \citep{gibson1998,demberg2008}, operationalized as Mean Dependency Distance \citep{liu2008}: since written English minimizes dependency length wherever grammatically possible \citep{temperley2007,gildea2015}, unusually long dependencies mark a departure from the baseline \citep[cf.][]{levy2008}. Finally, MMLU rankings are sensitive to semantically invariant perturbations such as prompt templates and option ordering \citep{alzahrani2024,pezeshkpour2024,zheng2024,mizrahi2024}, so Section~\ref{sec:selection} measures a noise floor inside our own population, and Appendix~\ref{sec:labels} refits the structural model against the 6.49\% label error rate of \citet{gema2025} and the contamination documented by \citet{sainz2023}, \citet{golchin2024} and \citet{deng2024}.

\section{Measurement Framework and Methodology}\label{measurement-framework-and-methodology}

\subsection{Data and Latent Parameter Calibration}\label{latent-parameter-calibration-via-item-response-theory}

We use the complete MMLU test set ($I = 14{,}042$ items across 57 subjects) and a model population of $J = 1{,}000$ open-weights models from the Hugging Face Open LLM Leaderboard \citep{beeching2023}, spanning the Deepseek, Falcon, Llama, Mistral, Pythia, Qwen, Solar, Yi and Mixture-of-Experts families, with parameter counts from 3.2 million to over 180 billion, so that the binary response matrix $\mathbf{Y} \in \{ 0,1\}^{J \times I}$ yields approximately 14 million observations. That population is a leaderboard rather than a designed sample and is substantially redundant, reducing to 869 distinct checkpoints and 50 base-family-and-size combinations, which is harmless for item parameters estimated from a thousand models but inflates any quantity counted over models, so every model-level statistic below is reported on the deduplicated population as well.

Item difficulty is estimated using the Two-Parameter Logistic model \citep{birnbaum1968,lord1968}, $P(Y_{ji} = 1 \mid \theta_{j},a_{i},b_{i}) = [ 1 + \exp( - a_{i}(\theta_{j} - b_{i}))]^{-1}$, where $b_{i}$ is the difficulty of item $i$ and $a_{i} > 0$ its discrimination, with priors $\theta_{j} \sim \mathcal{N}(0,1)$, $b_{i} \sim \mathcal{N}(0,1)$ and $a_{i} \sim \mathrm{HalfNormal}(1.0)$. With 1,000 models per item the likelihood dominates the prior. We estimate by Stochastic Variational Inference in NumPyro and JAX, not Joint Maximum Likelihood, which suffers from the incidental parameters problem \citep{neyman1948}: the nuisance parameters $\{\theta_{j}\}$ grow with $J$, and item estimates stay inconsistent as a result. SVI marginalizes over the ability distribution instead. Calibrated difficulty spans $b_{i} \in [ - 5.00,7.12]$ ($\sigma = 2.24$) and ability spans $\theta_{j} \in [ - 2.02,3.14]$ ($\sigma = 1.03$). Calibrating two independent random cohorts of 500 models separately, the two difficulty vectors correlate at $r_{hh} = 0.9921$ across all 14,042 items, a full-test reliability of approximately $0.9960$ under Spearman--Brown. The estimates are therefore mostly free of incidental measurement error from model sampling, and downstream disattenuation corrections are not needed.

\subsection{A Priori Domain Grouping}\label{a-priori-domain-grouping}

To prevent post hoc selection bias on observed response patterns or text features, we fix the subject partition before any path model is estimated, adopting the official taxonomy of MMLU's designers \citep{hendrycks2021}, which gives $I_{\text{STEM}} = 3{,}153$ items across 19 subjects against $I_{\text{non-STEM}} = 10{,}889$ items across 38 subjects. It is well established in educational psychology \citep{newcombe2009} that mathematics and science learning recruit partly distinct cognitive processes from verbal and social reasoning. That partition is a union of whole subjects, and because MMLU items were written in subject blocks, items within a subject share topic, source, authoring and format, so they are not independent observations. The effective sample size is closer to 57 than to 14,042, and Section~\ref{sec:invariance} accordingly treats the subject, not the item, as the unit of clustering and of resampling.

\subsection{The Five-Dimension, Nine-Indicator Framework}\label{the-five-dimension-nine-indicator-framework}

We operationalize item complexity as five separable dimensions extracted deterministically from the prompt text, because two items can share identical word counts and syntactic structures while differing in logical nesting, lexical abstractness or factual obscurity. Table~\ref{tab:indicators} gives each indicator's definition and its cognitive or linguistic grounding. Two construction details do not reduce to a table entry.

\begin{table*}[t]
\centering\scriptsize
\setlength{\tabcolsep}{5pt}
\begin{tabular}{@{}>{\raggedright\arraybackslash}p{0.235\textwidth}
                  >{\raggedright\arraybackslash}p{0.715\textwidth}@{}}
\toprule
\textbf{Indicator} & \textbf{Definition} \\
\midrule
\multicolumn{2}{@{}l}{\emph{Dimension 1: topological reasoning load.} Procedural knowledge in ACT-R \citep{anderson1983}} \\
WSCG Depth ($W_\mathrm{scg}$) & Weighted longest path $\max_{p \in \mathcal{P}}\sum_{v \in p} w(v)$ through the symbolic complexity DAG \\
WSCG Nodes ($N_\mathrm{scg}$) & Operational nodes in that DAG, i.e.\ distinct reasoning operations or leaves \\
\midrule
\multicolumn{2}{@{}l}{\emph{Dimension 2: syntactic complexity.} Dependency Locality Theory \citep{gibson1998}} \\
Syntactic MDD ($D_\mathrm{mdd}$) & $\frac{1}{n_{i}-1}\sum_{t=2}^{n_{i}}|t - h(t)|$ over the Stanza parse, with $h(t)$ the head index of token $t$ \citep{liu2008} \\
Syntactic Depth ($H_\mathrm{syn}$) & Longest leaf-to-root path in the dependency tree \\
\midrule
\multicolumn{2}{@{}l}{\emph{Dimension 3: epistemological specificity.} Declarative knowledge in ACT-R \citep{anderson1983}} \\
Entity Density ($D_\mathrm{ner}$) & $|E_\mathrm{ner}| / N_\mathrm{tokens}$, named-entity tokens from SpaCy NER as a fraction of all tokens \\
Zipf Rarity ($R_\mathrm{zipf}$) & $\frac{1}{|V_{z}|}\sum_{w \in V_{z}}(8.0 - Z(w))$ over content words, $Z$ the wordfreq Zipf frequency \citep{speer2022} \\
\midrule
\multicolumn{2}{@{}l}{\emph{Dimension 4: semantic complexity.} Syntax/semantics double dissociation \citep{berndt1996}} \\
Concreteness ($C_\mathrm{bry}$) & Mean of the \citet{brysbaert2014} 1--5 norms over covered content words, higher denoting more concrete \\
AMR Depth ($D_\mathrm{amr}$) & Longest path through the BART-AMR semantic DAG \citep{banarescu2013}, i.e.\ nesting with surface syntax stripped \\
\midrule
\multicolumn{2}{@{}l}{\emph{Dimension 5: adversarial logic structure.} Non-monotonic inference under negation} \\
Adversarial Negation ($\widetilde{S}_\mathrm{neg}$) & OLS residual, on token count, of raw negation scope counted over the dependency scope of a 59-term trigger lexicon \\
\bottomrule
\end{tabular}
\caption{Five dimensions and nine deterministic indicators of structural complexity, computed from the prompt stem.}
\label{tab:indicators}
\end{table*}

\textbf{The Weighted Symbolic Complexity Graph.}\label{sec:d1} Dimension~1 measures how much sequential multi-step reasoning an item requires, since retrieving two facts is a different computational demand from chaining them through a nested conditional and no length or readability measure separates the two. For each prompt we construct a node-weighted directed acyclic graph $G_{w} = (V,E,w)$ over the Stanza parse, restricting $V$ to operational tokens identified through a cognitive lexicon of 13 lexical categories plus a symbolic-operator class. Each vertex carries an intrinsic weight $w(v) \in [ 0.0,5.0]$ on eight active tiers matching its theoretical inferential friction: boolean conjunctions take $w = 1.0$, causative conditionals $w = 4.0$ and \LaTeX{} integrals $w = 5.0$. Edges are projected from grammatical dependency arcs and supplemented by state-tracking edges whenever an entity referenced in sentence $i$ reappears in sentence $j > i$. Because standard DAG path algorithms evaluate edge weights rather than node weights, WSCG Depth maps each target node's weight onto its incoming edge before solving for the longest path. The weight assignments are heuristically motivated, a vulnerability we test directly: perturbing the tiers by $\pm 1.0$ over 100 Monte Carlo draws, one tier at a time, leaves all three primary conclusions intact across all 123 schemes tested (Appendix~\ref{sec:sensitivity}). The highest tier ($w = 5.0$, for integral and nabla operators) is inert here, because MMLU mathematical content is overwhelmingly ASCII. That leaves Dimension~1 blind to one component of STEM difficulty, a scope condition we return to in Section~\ref{limitations}.

\textbf{Sensitivity to prompt length and to the answer options.} Because nouns occupy the lowest active weight tier, one might object that topological reasoning load is merely correlated with prompt length. This holds for WSCG Nodes, which correlates with token count at $r = 0.985$, but fails for WSCG Depth ($r = 0.580$), and replacing both Dimension~1 indicators with their OLS residuals on token count preserves the stratified fits ($W(9) = 134.79$), as does clamping the baseline $0.5$ tier to zero. Separately, all nine indicators compute complexity from the question stem alone. Eight option-set covariates yield complementary predictive power without eliminating the cross-domain contrast (Appendix~\ref{sec:form}), and we restrict the primary analysis to the stem because its dimensions isolate theoretically grounded cognitive demand.

\section{Results}\label{results}

\subsection{Global Path Analysis and Baselines}\label{sec:global}

We begin with a pooled OLS regression across all 14,042 items, regressing IRT difficulty on the nine-indicator framework as $b_{i} = \beta_{0} + \sum_{k = 1}^{9}\beta_{k}X_{ki} + \epsilon_{i}$, with HC3 heteroskedasticity-consistent standard errors \citep{mackinnon1985}. Section~\ref{sec:invariance} replaces these with subject-clustered covariances for inferences that carry weight. Eight predictors reach significance and the omnibus test is extreme, yet the model explains only $3.57\%$ of global variance. That is not a multicollinearity artifact: variance inflation factors stay in the $1.025$--$2.379$ range. A figure this size is nonetheless what should be expected here, not a sign the framework has failed. Educational measurement obtains $R^{2}$ in the $0.10$--$0.30$ band from cognitive text features only on \emph{homogeneous} item pools \citep{embretson1987}, and MMLU is 57 of them stacked together. The framework also isolates only the \emph{structural} component of difficulty. Everything content-specific is left on the table by design. And because $77.55\%$ of items are non-STEM, the pooled coefficients are dominated by that majority regime, which dilutes a signal concentrated in STEM to near-invisibility.

\begin{table}[t]
\centering\scriptsize
\setlength{\tabcolsep}{3.5pt}
\begin{tabular}{@{}lrrr@{}}
\toprule
\textbf{Predictor set} & \textbf{Global} & \textbf{STEM} & \textbf{n-STEM} \\
\midrule
Word count                  & $0.0059$ & $0.0123$ & $0.0133$ \\
$+$ Flesch--Kincaid         & $0.0086$ & $0.0289$ & $0.0143$ \\
Nine indicators             & $\mathbf{0.0331}$ & $\mathbf{0.0732}$ & $\mathbf{0.0361}$ \\
Indicators $+$ surface      & $0.0383$ & $0.0844$ & $0.0360$ \\
Option-set indicators       & $0.0684$ & $0.0825$ & $0.0559$ \\
Indicators $+$ options      & $0.0808$ & $0.1196$ & $0.0713$ \\
\midrule
Subject identity only       & $0.1354$ & $0.1246$ & $0.1197$ \\
\emph{Encoder ceiling}      & $0.1269$ & $0.1330$ & $0.1113$ \\
\emph{$+$ subject identity} & $0.1596$ & $0.1584$ & $0.1431$ \\
\bottomrule
\end{tabular}
\caption{Cross-validated $R^2$ (5-fold mean). The ceiling is a ridge regression on a frozen MiniLM encoder (stem $+$ options).}
\label{tab:baselines}
\end{table}

The relevant question is what that figure should be compared against, and Table~\ref{tab:baselines} answers it in two ways. Sheer length predicts difficulty weakly and a readability index adds a little more, chiefly in STEM. Against that baseline the framework captures $3.8 \times$ the cross-validated variance globally and $2.5 \times$ in STEM, significantly in every partition (globally $F(9,14029) = 51.15$, $p = 6.4 \times 10^{-92}$). The second comparison is a dense sentence encoder reading the whole item, which estimates the ceiling on text-predictable difficulty. That ceiling is itself low, at $0.1269$ pooled and $0.1330$ in STEM: most MMLU item difficulty simply is not recoverable from item text, by any method. Measured against it, the nine indicators recover $56.5\%$ of text-predictable difficulty in STEM and $32.6\%$ outside it, and the option-set indicators take the STEM figure to $89.9\%$. Subject identity alone reaches $0.1354$ pooled. A good deal of what an encoder knows about an item, in other words, is really just which subject it came from, whereas the nine indicators carry no subject label at all. The framework against the ceiling is therefore the comparison that matters.

\begin{table*}[t]
\centering\scriptsize
\setlength{\tabcolsep}{3.5pt}
\begin{tabular}{lrrrrrrrr}
\toprule
& \textbf{Pooled} & \multicolumn{2}{c}{\textbf{STEM}} & \multicolumn{2}{c}{\textbf{non-STEM}} & \multicolumn{3}{c}{\textbf{Contrast}} \\
\cmidrule(lr){2-2}\cmidrule(lr){3-4}\cmidrule(lr){5-6}\cmidrule(lr){7-9}
\textbf{Complexity Indicator} & \textbf{$b$} & \textbf{$b$} & \textbf{$\beta$} & \textbf{$b$} & \textbf{$\beta$} & \textbf{$Z_{\text{HC3}}$} & \textbf{$t_{\text{CR2}}$} & \textbf{$p_{\text{boot}}$} \\
\midrule
WSCG Depth ($W_\mathrm{scg}$)                      & $0.0696^{***}$  & $0.0602$  & $0.073$  & $0.0620$  & $0.115$  & $-0.08$ & $-0.05$ & $1.00$ \\
WSCG Nodes ($N_\mathrm{scg}$)                      & $-0.0023^{*}$   & $0.0189$  & $0.088$  & $-0.0007$ & $-0.010$ & $2.93$  & $1.76$  & $0.46$ \\
Syntactic MDD ($D_\mathrm{mdd}$)                   & $0.1654^{***}$  & $0.3694$  & $0.102$  & $0.0146$  & $0.004$  & $4.05$  & $2.68$  & $0.12$ \\
Syntactic Depth ($H_\mathrm{syn}$)                 & $0.0027$        & $-0.0584$ & $-0.037$ & $0.0130$  & $0.012$  & $-1.81$ & $-1.28$ & $0.84$ \\
Zipf Rarity ($R_\mathrm{zipf}$)                    & $0.0971^{*}$    & $0.2523$  & $0.047$  & $-0.0521$ & $-0.010$ & $2.90$  & $1.85$  & $0.46$ \\
Entity Density ($D_\mathrm{ner}$)                  & $0.0108^{***}$  & $0.0356$  & $0.140$  & $0.0002$  & $0.001$  & $7.03$  & $3.41$  & $\mathbf{0.014}^{\dagger}$ \\
Concreteness ($C_\mathrm{bry}$)                    & $0.4127^{***}$  & $-0.1642$ & $-0.021$ & $0.5883$  & $0.078$  & $-4.94$ & $-1.93$ & $0.46$ \\
AMR Depth ($D_\mathrm{amr}$)                       & $0.0516^{***}$  & $0.0523$  & $0.035$  & $0.0608$  & $0.055$  & $-0.28$ & $-0.14$ & $1.00$ \\
Adversarial Negation ($\widetilde{S}_\mathrm{neg}$)& $-0.0232^{***}$ & $-0.0343$ & $-0.033$ & $-0.0177$ & $-0.038$ & $-0.84$ & $-0.82$ & $1.00$ \\
\bottomrule
\end{tabular}
\caption{OLS path coefficients for item difficulty ($b$: unstandardized; $\beta$: within-partition standardized). Contrasts test cross-domain homogeneity under progressively stricter assumptions: independent items ($Z_{HC3}$), 57-subject cluster-robust covariances ($t_{CR2}$), and a 9,999-replicate wild cluster bootstrap with Holm correction ($p_{boot}$, with $\dagger$ marking the sole survivor). $^{***}p<0.001$, $^{*}p<0.05$.}
\label{tab:stratified}
\end{table*}

\subsection{Domain-Stratified Bifurcation}\label{domain-stratified-bifurcation}

To recover the signal suppressed by pooling, we estimate separate OLS regressions on each partition (Table~\ref{tab:stratified}). Structural complexity explains roughly twice the cross-validated difficulty variance in STEM ($0.0732$) as outside it ($0.0361$), a ratio robust to the outcome's scale, since in logit-squared units the indicators account for $2.46\times$ more variance in STEM (bootstrap CI $[1.80,3.36]$). However, because MMLU is clustered by subject, arbitrary 19-subject groupings routinely produce $R^{2}$ gaps of this size ($p = 0.08$, Appendix~\ref{sec:clustered}). Our structural claim therefore rests on slope contrasts, not on fit magnitude. We checked this: standardizing within partitions leaves the identical five contrasts surviving Holm correction.

Entity Density ($D_{\text{ner}}$) shows the most robust divergence, surviving all corrections ($t_{\text{CR2}} = 3.41$, $p_{\text{Holm}} = 0.011$, wild cluster bootstrap $p = 0.014$). In STEM, named entities act as operational constraints compounding the reasoning chain ($\beta = 0.140$). Outside STEM the coefficient is indistinguishable from zero ($b = 0.0002$), because entities there mostly supply associative context that facilitates recall instead of adding computational load. Over their shared range the STEM relationship rises monotonically while the non-STEM fit traces a non-monotone curve (Appendix~\ref{sec:form}). Syntactic MDD ($D_{\text{mdd}}$) shows a similar split: a strong positive coefficient in STEM ($b = 0.3694$), near-zero outside it ($b = 0.0146$), clearing clustered inference ($t_{\text{CR2}} = 2.68$, bootstrap $p = 0.015$) though not Holm correction. Dependency Locality Theory's processing-cost predictions hold for procedural computation, in other words, but not for factual recall. Lexical Concreteness ($C_{\text{bry}}$) reverses sign across the boundary ($Z_{\text{HC3}} = -4.94$), but that reversal does not survive subject clustering ($t_{\text{CR2}} = -1.93$), and a restricted cubic spline shows the non-STEM relationship to be significantly non-linear (Appendix~\ref{sec:form}). Four indicators do not differ under any estimator. Across the full set, five contrasts clear Holm correction under HC3, two under CR1, one under the bootstrap: the bifurcation is specific, not global.

Individual subject-level fits are descriptive only, since nine predictors on 100-item pools give an 11:1 observation-to-parameter ratio and severe in-sample optimism, and partition-level effect sizes are small (Cohen's $q = 0.0918$, CI $[0.052,0.136]$, \citealp{cohen1988}). The supportable claim is therefore not that domains differ in \emph{how much} variance structure explains, but in \emph{which} structural features do the explaining.

\subsection{Testing Structural Homogeneity}\label{sec:invariance}

We now test whether the coefficient differences in Table~\ref{tab:stratified} constitute a significant violation of structural homogeneity, regressing difficulty on the nine indicators, a STEM dummy and their interactions, where homogeneity is the null that all nine interaction terms are zero. Assuming item independence, the HC3 Wald test decisively rejects it ($W(9) = 138.36$, $p = 2.3 \times 10^{-25}$). We exclude the intercept interaction ($W(10) = 375.76$), which reflects a baseline difficulty shift rather than a mapping difference.

Because MMLU items are clustered within 57 subjects, HC3 covariances underestimate standard errors \citep{liang1986}. The Bell--McCaffrey CR2 correction \citep{bell2002} inflates slope standard errors by a median factor of $1.65$, reducing the joint statistic to $W(9) = 28.21$ ($p = 8.8 \times 10^{-4}$, and $W(9) = 33.13$, $p = 1.3 \times 10^{-4}$ under CR1), which still rejects against $\chi_{9}^{2}$. Because 57 is a small number of clusters we further apply a wild cluster bootstrap with the null imposed \citep{cameron2008}, which yields an anti-conservative reference distribution (95th percentile $34.7$ against nominal $16.9$) and renders the joint test marginal ($p = 0.060$). Since a 9-degree-of-freedom test dilutes power across contrasts already shown to be null, bootstrapping one contrast at a time proves decisive for Entity Density ($p_{\text{Holm}} = 0.014$) and suggestive for Syntactic MDD ($p = 0.015$ uncorrected). We report the nested likelihood ratio test ($\Lambda(11) = 454.86$) for comparability with the invariance literature \citep{kline2016,wilks1938} but rely on the resampling tests, since the residuals exhibit significant skew ($1.37$) and heteroskedasticity (Breusch--Pagan $130.70$).

Permutation tests confirm the necessity of cluster-aware inference. An item-level permutation of the STEM label yields a maximum $R^{2}$ gap of $0.0315$, below the observed $0.0412$ ($p < 10^{-4}$), but that null ignores between-subject heterogeneity: permuting whole 19-subject blocks widens it by a factor of $3.2$ and renders the gap non-significant ($p = 0.08$), while the slope-sensitive likelihood-ratio statistic still clears its block null ($p = 0.049$). Whether structural complexity predicts difficulty \emph{better} in STEM is therefore sensitive to cluster structure, but whether it predicts difficulty \emph{differently} is robust. Independent of these standard errors, the confirmatory two-dimensional IRT model (Appendix~\ref{sec:mirt}) yields a latent correlation of $\rho = 0.965$ with no posterior mass above $0.99$, improves held-out prediction by $+243.6$ nats on 5\% masked response cells (paired $z = 8.2$), and is corroborated by a model-free matched-length subscale correlation of $0.977$ over 200 random splits ($[0.974,0.980]$), with no replicate reaching unity.

Structural homogeneity is therefore rejected: the mapping from text-level complexity to IRT-estimated difficulty is not constant across partitions, an item-side analogue to measurement non-equivalence \citep{meredith1993}. This licenses claims about \emph{which} structural properties carry difficulty in each domain, not that one domain affords more predictable difficulty overall. The conclusion survives a Rasch refit (Appendix~\ref{sec:sensitivity}), holds on each predictor's common support controlling for option-set indicators (Appendix~\ref{sec:form}), and subsists after removing every item flagged as erroneous by MMLU-Redux (Appendix~\ref{sec:labels}).

\section{Consequences for Leaderboard Validity}\label{consequences-for-leaderboard-validity}

If the complexity-to-difficulty mapping differs structurally across domains, the interpretation of the aggregate as a unified ability measure is undermined. Because $77.55\%$ of the items are non-STEM, the aggregate is arithmetically weighted toward non-STEM performance to begin with. What turns that weighting from a counting artifact into a measurement asymmetry is the non-invariance established in Section~\ref{sec:invariance}. The next two subsections work through what that asymmetry costs in practice.

\subsection{Reasoning Stability}\label{sec:inversion}

\begin{figure*}[t]
\centering
\includegraphics[width=0.88\textwidth]{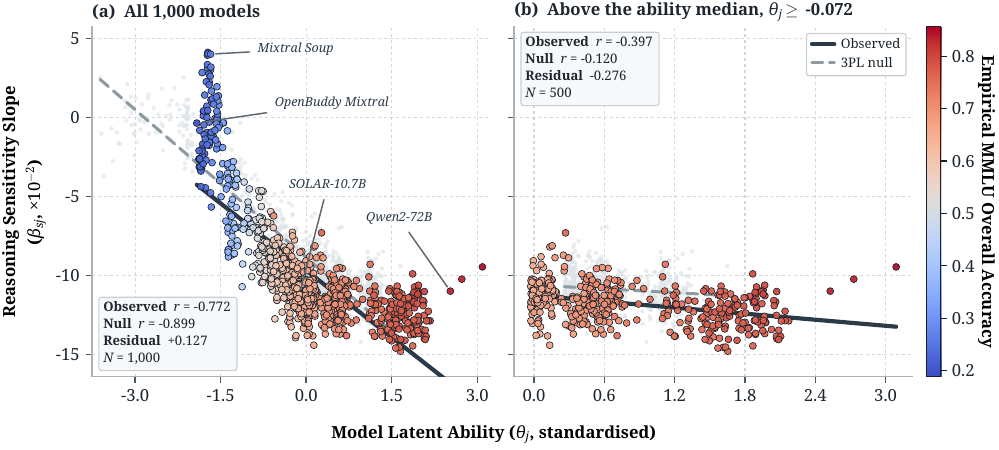}
\caption{Reasoning sensitivity slopes ($\beta_{sj}$) against latent ability ($\theta_j$): (a) all 1,000 models, (b) the upper-ability half. Filled/solid: observed. Open/dashed: 3PL nulls without ability-by-complexity interaction.}
\label{fig:inversion}
\end{figure*}

A Reasoning Sensitivity Coefficient ($\beta_{sj}$) can be estimated per model by logistic regression of binary accuracy on WSCG Depth. Restricted to STEM and fitted in log-odds space, $\beta_{sj}$ correlates with latent ability at $r = -0.772$ across all 1,000 models, and at $-0.512$ and $-0.397$ once the bottom quartile and bottom half are excluded, suggesting that higher-ability models degrade more steeply as reasoning depth increases. That two-stage estimate turns out to be severely confounded, though. Success probability is bounded at $c = 0.25$, so empirical log-odds are mechanically compressed for weaker models, and their slopes flatten relative to stronger models' even if no true ability-by-complexity interaction exists at all. Simulating response matrices from a 3PL model ($c = 0.25$) fitted to the STEM submatrix and passing them through the identical procedure isolates that geometric artifact: the generative null yields correlations of $-0.899$, $-0.355$ and $-0.119$ across the three samples, accounting for 116\%, 69\% and 30\% of the observed values while preserving construct consistency (Figure~\ref{fig:inversion}), its abilities and difficulties correlating at $0.907$ and $0.892$ with the 2PL baseline.

To handle the guessing floor natively and avoid interpreting two-stage residuals, we replace this procedure by estimating the interaction directly inside the response model,
\[\eta_{ji} = a_{i}( \theta_{j} - b_{i} ) + ( \gamma_{0} + \gamma_{1}\theta_{j} )W_{i},\]
with $P( Y_{ji} = 1 ) = c + (1 - c)\sigma( \eta_{ji} )$, $c = 0.25$ and $W_{i}$ standardized, so that $\gamma_{1}$ captures the interaction as a native parameter. Structural complexity depresses success at mean ability ($\gamma_{0} = -0.300$, CI $[-0.326,-0.272]$), and the interaction remains negative but highly modest, at $\gamma_{1} = -0.086$ logits per standard deviation of WSCG Depth per unit of ability (CI $[-0.093,-0.079]$). Higher-ability models \emph{do} degrade more steeply under increased reasoning depth. The true effect, though, is much smaller than the naive correlation implies. Read the sign and magnitude of the point estimate as the takeaway rather than the interval itself, since mean-field posteriors understate spread. The effect, also, is estimated strictly on the STEM partition.

\subsection{Consequences for Downstream Model Selection}\label{sec:selection}

In construct validity terms \citep{cronbach1955,messick1989}, the MMLU aggregate conflates Retrieval Capacity and Reasoning Stability, weighting the former more heavily, and practically this affects rank stability. Weighted Kendall's $\tau$ \citep{shieh1998,vigna2015}, which prioritizes frontier rank agreement, is higher for non-STEM ($\tau_{w} = 0.9887$) than for STEM ($0.9634$). Subsampling non-STEM to STEM's 3,153 items over 200 draws halves the gap to a matched $0.9756$ (CI $[0.9633,0.9827]$), and the STEM value sits below 194 of these 200 draws ($p = 0.035$), confirming a real, though reduced, one-sided asymmetry.

The asymmetry manifests directly in model selection. Of the Top 50 aggregate models, 92\% retain their status on non-STEM but only 78\% do so on STEM, and for the Top 10 retention is 100\% against 80\%, so relying on the aggregate leaderboard for a STEM-focused application carries a 22\% displacement rate (bootstrap CI $[16\%,28\%]$). This displacement is largely driven by the smaller STEM item sample, since a size-matched random-item null yields 14.4\% ($p = 0.086$) and a subject-block matched null 22.4\% ($p = 0.61$). Deduplicating to the 869 base models raises it to 28.8\%. We treat the 22\% figure as a quantified cost to practitioners, not as independent evidence for construct separation on its own. For an empirical noise floor \citep{alzahrani2024}: seven duplicate checkpoint pairs agreeing within 0.1\% aggregate accuracy still disagree by a median of 1 aggregate and 3 STEM ranks (maxima 11 and 23), which puts shifts below roughly ten places in the range of ordinary noise. That noise is not one-directional, at least. Standardizing both abilities of the two-dimensional model, the $\theta_{\text{STEM}} - \theta_{\text{non-STEM}}$ gap has a standard deviation of $0.19$, and while 83.5\% of models sit within $0.25\sigma$ of the identity line, the positive tail is systematic and belongs to reasoning-tuned models such as Qwen2-72B ($+0.88$). The aggregate summarizes the general population adequately while being unrepresentative of a minority optimized for procedural reasoning.

\section{Discussion}\label{discussion-1}

\subsection{Implications for Benchmark Design}\label{implications-for-benchmark-design}

The heuristic reforms of MMLU-Pro \citep{wang2024} are useful but theoretically insufficient, since filtering by difficulty or discrimination removes uninformative items and leaves the non-invariance of the mapping untouched, whereas our deterministic framework supplies the missing structural audit without the stochastic variance and API-dependence of LLM-as-annotator methods \citep{gilardi2023}. We recommend that future benchmarks report distinct retrieval and reasoning capacity indices alongside aggregate scores. Unlike category subscores, these are computable \emph{a priori} from item text before any model is evaluated, are explanatory in distinguishing whether a domain is difficult because it demands deep inference or because it presupposes rare facts, and are independent of designer taxonomies, which routinely misalign with structural reality: MMLU's \emph{Business Ethics} and \emph{High School Macroeconomics} are highly structurally sensitive, yet labeled non-STEM. Applying the framework to MMLU-Pro will test whether its construction resolves structural heterogeneity or merely displaces it upward.

\subsection{Limitations}\label{limitations}

\textbf{Variance explained and scope.} As the dense encoder ceiling establishes (Section~\ref{sec:global}), structural complexity is a significant but not dominant determinant of difficulty. Our instrument reads text only. Because MMLU's mathematical content is overwhelmingly ASCII, the top symbolic weight tier stays inert, and Zipf rarity falls back to the population mean for items with no scorable content words. That leaves the instrument weakest exactly where the abstract STEM signal is strongest. This evaluation also targets open-weights models, and whether these sensitivities extend to closed frontier models remains open.

\textbf{Statistical and methodological bounds.} Several findings are marginal against their strictest respective controls: the between-domain $R^{2}$ gap against a subject-block permutation ($p = 0.08$), the Top-50 displacement rate against a size-matched control ($p = 0.09$), and the joint homogeneity test under a wild cluster bootstrap ($p = 0.060$). We adjust for effective sample size by modeling 57 subject clusters, not 14,042 items, and 50 base-model families, not 1,000 checkpoints. Individual subject-level inferences end up underpowered as a result. The confounds we control explicitly, namely prompt length, answer-option structure, label error and pretraining contamination (Appendices~\ref{sec:form} and \ref{sec:labels}), are bounded by this, not eliminated. The ideal specification would be a fully explanatory IRT model propagating difficulty uncertainty into the domain contrast while natively absorbing subject clustering. We were unable to make that model identify, and Appendix~\ref{sec:mirt} documents the attempt as a formal negative result.

\textbf{Construct invariance limits.} The rejection of structural homogeneity is an \emph{item-side} phenomenon and does not formally license \emph{person-side} measurement non-equivalence \citep{meredith1993}. Our design does not strictly prove a single latent $\theta$ incomparable across models. Mantel--Haenszel tests do reveal differential item functioning (Appendix~\ref{sec:labels}), but that is a supplementary finding rather than a corollary of the structural mapping shift.

\section{Conclusion}\label{conclusion}

IRT calibration, multi-group analysis and confirmatory two-dimensional IRT applied to 14 million response observations from 1,000 language models demonstrate that MMLU's partitions do not measure a single interchangeable ability: the latent correlation between STEM and non-STEM ability is strictly below unity at $\rho = 0.965$, the disattenuated subscale correlation at matched test length is $0.977$, and the two-dimensional model improves held-out prediction. Crucially, the mapping from text structure to item difficulty is structurally non-invariant across partitions, a rejection holding under item-level inference ($W(9) = 138.36$) and subject-clustered covariances ($W(9) = 28.21$ under CR2), and the divergence is most absolute for named entity density, which acts as an operational constraint compounding computational load in STEM items but as an associative retrieval cue elsewhere. The heterogeneity carries downstream consequences: Top-50 aggregate selection displaces 22\% of STEM-appropriate choices, and once the guessing floor is controlled natively, higher-ability models degrade more steeply under procedural reasoning depth ($\gamma_{1} = -0.086$). Because the MMLU aggregate weights declarative retrieval capacity over procedural reasoning stability, it inherently favors models optimized for retrieval. We release the framework so that future benchmarks can be structurally audited before they are adopted as unified capability measures.

\bibliography{refs}
\bibliographystyle{acl_natbib}

\appendix
\setcounter{table}{0}\setcounter{figure}{0}
\renewcommand{\thetable}{A\arabic{table}}
\renewcommand{\thefigure}{A\arabic{figure}}
\makeatletter\@ifundefined{theHtable}{}{%
  \renewcommand{\theHtable}{A\arabic{table}}%
  \renewcommand{\theHfigure}{A\arabic{figure}}}\makeatother

\section{Sensitivity, Diagnostics and Clustering}\label{sec:sensitivity}

\textbf{Weight sensitivity.} We apply 123 weight tier perturbation schemes clamped to $[0.0,5.0]$: uniform shifts, single-tier shifts and 100 Monte Carlo draws from $[-1,+1]$. All primary conclusions hold across all 123 schemes, with structural homogeneity universally rejected (maximum $p = 1.26 \times 10^{-77}$), STEM $R^{2}$ strictly dominating non-STEM $R^{2}$ ($0.077$--$0.083$ against $0.031$--$0.043$) and the Syntactic MDD divergence remaining positive ($p < 0.01$, $Z \in [3.22,4.54]$).

\textbf{Diagnostics and robustness.}\label{sec:robustness} Partition-level fits are stable under 20 iterations of 5-fold cross-validation \citep{stone1974}, at $R_{cv}^{2}$ of $0.0732$ and $0.0361$ against in-sample $0.0792$ and $0.0380$, though individual subject pools show the standard small-$N$ failure mode \citep{babyak2004,yarkoni2017}. Residuals are right-skewed ($1.37$, $0.89$, $1.53$), Jarque--Bera rejects normality, and Breusch--Pagan \citep{breusch1979} and White \citep{white1980} tests reject homoskedasticity (LM $130.70$, $42.05$, $120.84$), which necessitates resampling rather than normal-theory inference. Refitting difficulty via a Rasch model \citep{rasch1960} recovers estimates correlating at $r = 0.894$ with the 2PL baseline and strengthens the contrast ($0.1038$ against $0.0676$, $W(9) = 167.21$). Filtering uninformative items ($a_{i} < 0.2$), which disproportionately affects STEM (31.9\% of removals against a 22.5\% pool share), raises STEM $R^{2}$ to $0.0862$ while leaving non-STEM invariant ($W(9) = 149.58$), and slope equality is rejected at every tested threshold.

\textbf{Clustering and size-matched nulls.}\label{sec:clustered} Predictor variances differ across partitions (WSCG Nodes has a $2.82\times$ larger standard deviation outside STEM), but standardizing within partitions leaves the identical five contrasts clearing Holm correction without sign reversals. Permuting whole 19-subject blocks rather than items widens the $R^{2}$-gap null more than threefold, to mean $0.0076$ (max $0.1056$, $p = 0.080$), and the block null for rank displacement at $K = 50$ is 22.4\% ($p = 0.609$) against a random-item null of 14.4\% ($p = 0.086$). Evaluated one contrast at a time under the wild cluster bootstrap, only Entity Density survives Holm correction ($p = 0.0015$ raw, $0.014$ corrected), with Syntactic MDD next at $0.015$ raw and $0.12$ corrected.

\section{Functional Form and Option Structure}\label{sec:form}

\textbf{Common support and functional form.} Restricting partitions to their intersected 1st--99th percentile ranges retains 97--99\% of items and preserves all conclusions: Syntactic MDD shifts to $0.4083/0.0168$ ($Z = 3.90$), Entity Density to $0.0353/0.0036$ ($Z = 5.49$) and Concreteness to $-0.1479/0.7601$ ($Z = -5.13$). A four-knot restricted cubic spline \citep{harrell2015} on the focal indicator significantly improves fit outside STEM for Syntactic MDD ($F(2,10877) = 17.43$) and Concreteness ($F = 15.86$) but not inside it ($p > 0.15$): Syntactic MDD and Entity Density rise monotonically in STEM but trace non-monotone curves outside it.

\textbf{Option-set indicators.}\label{sec:options} We extract eight choice-based indicators, for instance distractor homogeneity, option length dispersion and numeric proximity, using a frozen MiniLM sentence encoder \citep{reimers2019}. These reach $R_{cv}^{2} = 0.0684$ globally against $0.0338$ for the stem indicators, and the two families combined reach $0.0808$ globally and $0.1196$ in STEM. Refitting the homogeneity test on the 17-indicator set remains decisive ($W(17) = 138.86$), and the option indicators are themselves non-invariant ($W(8) = 51.97$). Their two strongest members are the key's stem-similarity advantage ($b = -1.477$) and maximum distractor similarity ($b = 1.056$): difficulty scales in direct proportion to how closely an item's best distractor imitates its key.

\section{Confirmatory Two-Dimensional IRT}\label{sec:mirt}

Each model receives a STEM ability and a non-STEM ability drawn from a bivariate normal with a free correlation, each item loads only on the dimension its designer-fixed subject assigns, and the model is estimated by SVI as in Section~\ref{latent-parameter-calibration-via-item-response-theory}. The latent correlation is $0.965$ with no posterior mass above $0.99$, and the variational objective improves over the unidimensional model. Because an improvement in a bound on in-sample fit is weak evidence when the larger model carries a thousand additional parameters, we also mask a random $5\%$ of response cells before estimation and score each model on cells it never observed: the two-dimensional model is better there by $+243.6$ nats, with a paired $z$ of $8.2$ over $703{,}696$ held-out cells. Disattenuating matched-length subscale scores by their Spearman--Brown reliabilities gives $0.977$ over 200 random splits at $1{,}576$ items per side ($[0.974,0.980]$), and because mean-field posteriors understate spread we rely on that split-half estimate for uncertainty.

We also fitted the natural companion analysis, an explanatory IRT model writing difficulty as a function of the indicators inside the response model with domain-varying slopes and a subject random effect \citep{fischer1973,deboeck2004}, but do not report its estimates, because they are not trustworthy: the fitted residual item scale ($4.93$ logits) exceeds the standard deviation of the difficulty distribution itself ($2.24$), the credible intervals are an order of magnitude narrower than the corresponding robust standard errors, and refitting at increasing step budgets reverses the sign of two of the nine domain interactions. The model is weakly identified as written, since a free per-item residual over $14{,}042$ items absorbs what the indicators would otherwise explain.

\section{Label Errors, Contamination and Transfer}\label{sec:labels}

\textbf{Label error and contamination.} Matching the \citet{gema2025} re-annotation to our pool covers 5,984 items and flags 374 (6.25\%), at a lower rate in STEM (4.70\%) than outside (6.98\%). As expected for incorrect keys, flagged items are significantly harder (mean $b = +1.86$ against $-0.10$, $t = 12.08$), but they do not drive the domain contrast: dropping them yields STEM and non-STEM $R^{2}$ of $0.0852$ and $0.0442$ ($W(9) = 140.71$), refitting on the verified subset alone yields $0.0934$ and $0.0141$ ($W(9) = 84.45$), and leaving the annotated subsample unfiltered yields $0.0828$ and $0.0115$ ($W(9) = 82.79$). The divergence lives in the subsample itself, not in how it was filtered. Recalibrating difficulty on an early cohort (99 models: Pythia, Falcon, Llama-2) and a recent one (606 models: Llama-3, Qwen2, Phi-3, Gemma, Mixtral, Yi-1.5) yields difficulty vectors correlated at $r = 0.832$, 86\% of the attainable $r = 0.967$ ceiling, and the contrast persists in both, at $0.0712/0.0505$ ($W(9) = 124.14$) and $0.0839/0.0577$ ($W(9) = 135.81$). Under the strictest control, flagged items removed and difficulty estimated on the early cohort alone, fits are $0.0751$ and $0.0552$.

\textbf{Respondent-side invariance and transfer.} Mantel--Haenszel DIF \citep{holland1988,dorans1993}, stratifying models into ability deciles with Llama ($n = 319$) as reference, classifies 42.9\% of items as B or C for Mixtral, 44.9\% for Solar and 30.2\% for Yi, and the proportion flagged is higher in STEM across all three families by 7.5, 7.5 and 7.2 points (all $z > 7.4$), an excess reproduced by a purified second pass over the 4,714 items classified A throughout. Transferring the stem-only framework to HellaSwag \citep{zellers2019} returns a clean null across the designer-fixed ActivityNet/WikiHow boundary ($W(8) = 3.64$, $p = 0.89$), whereas the MMLU control refitted on the identical 219 models reproduces the bifurcation ($W(8) = 150.48$) and six ending-set indicators reject source invariance on those same HellaSwag items ($W(6) = 65.33$), so the test does not mechanically fire on arbitrary large partitions. On GSM8K \citep{cobbe2021} the WSCG tracks annotated calculator-step counts beyond prompt length (partial $r = 0.155$, $p = 1.9 \times 10^{-8}$), whereas on MATH \citep{hendrycks2021math} it fails to track human-assigned difficulty once length is controlled. Deduplicated re-runs across the 869 checkpoints reproduce the ability-to-sensitivity correlations to within $0.002$. Redundancy is inflating significance here, not the effect size itself.

\end{document}